\documentclass[10pt,twocolumn,letterpaper]{article}

\usepackage{cvpr}
\usepackage{times}
\usepackage{epsfig}
\usepackage{graphicx}
\usepackage{amsmath}
\usepackage{amssymb}
\usepackage{bbm}
\usepackage{multirow}
\usepackage{float}
\usepackage[pagebackref=true,breaklinks=true,colorlinks,bookmarks=false]{hyperref}

\cvprfinalcopy % *** Uncomment this line for the final submission

\begin{document}

%%%%%%%%% TITLE
\title{Generator Pyramid for High-Resolution Image Inpainting}

\author{Leilei Cao, Tong Yang\\
OPPO Research\\
\and
Yixu Wang\\
Johns Hopkins University\\
\and
Bo Yan, Yandong Guo\\
OPPO Research \\
}

\maketitle
%\maketitle
%\thispagestyle{empty}

%%%%%%%%% ABSTRACT
\begin{abstract}
   Inpainting high-resolution images with large holes challenges existing deep learning based image inpainting methods. We present a novel framework---PyramidFill for high-resolution image inpainting task, which explicitly disentangles content completion and texture synthesis. PyramidFill attempts to complete the content of unknown regions in a lower-resolution image, and synthesis the textures of unknown regions in a higher-resolution image, progressively. Thus, our model consists of a pyramid of fully convolutional GANs, wherein the content GAN is responsible for completing contents in the lowest-resolution masked image, and each texture GAN is responsible for synthesizing textures in a higher-resolution image. Since completing contents and synthesising textures demand different abilities from generators, we customize different architectures for the content GAN and texture GAN. Experiments on multiple datasets including CelebA-HQ, Places2 and a new natural scenery dataset (NSHQ) with different resolutions demonstrate that PyramidFill generates higher-quality inpainting results than the state-of-the-art methods. To better assess high-resolution image inpainting methods, we will release NSHQ, high-quality natural scenery images with high-resolution 1920$\times$1080.
\end{abstract}

%%%%%%%%% BODY TEXT
\section{Introduction}
Image inpainting, as a fundamental low-level vision task, has attracted much attention from academic and industry. A wide range of vision and graphics applications refer to image inpainting, e.g., object removal\cite{Criminisi, PatchMatch}, image restoration\cite{Du2020}, manipulation\cite{Pan2020}, and super-resolution\cite{Yang2020}. Image inpainting aims to synthesize alternative contents in missing regions of an image, which is visually realistic and semantically correct\cite{DeepFillV1}. 

\begin{figure}[t]
\centering
\begin{tabular}
	{@{\hspace{0.0mm}}c@{\hspace{0.5mm}}c@{\hspace{0.5mm}}c@{\hspace{0.5mm}}c@{\hspace{0.0mm}}}
   	\includegraphics[width=0.115\textwidth]{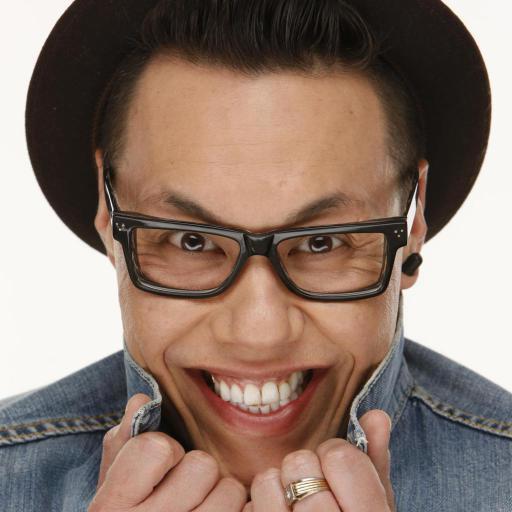}&
		\includegraphics[width=0.115\textwidth]{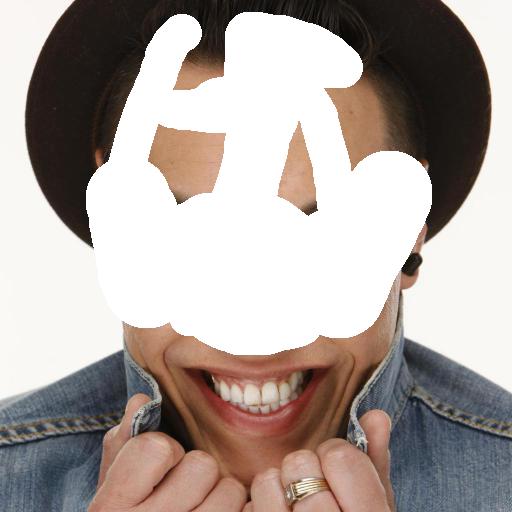}&
		\includegraphics[width=0.115\textwidth]{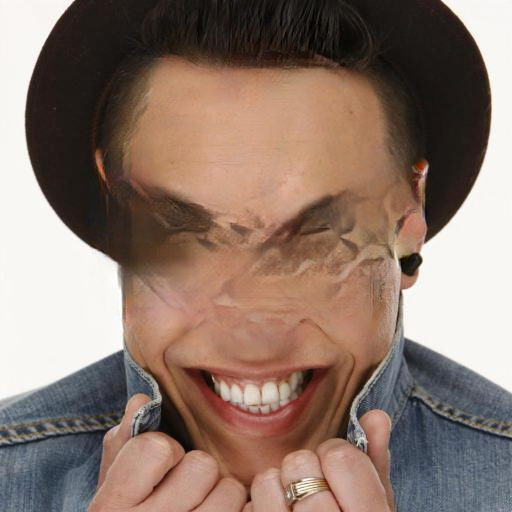}&
		\includegraphics[width=0.115\textwidth]{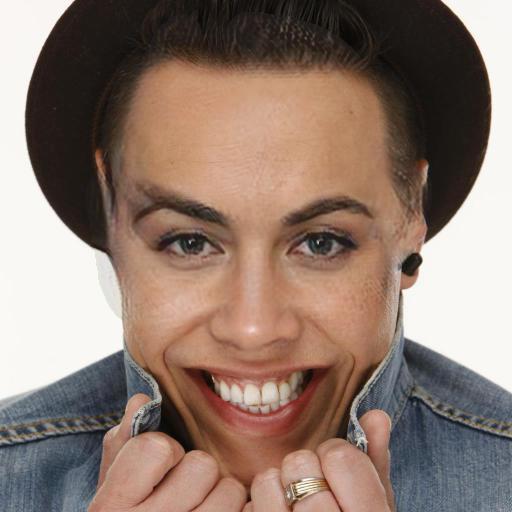}\\
		\includegraphics[width=0.115\textwidth]{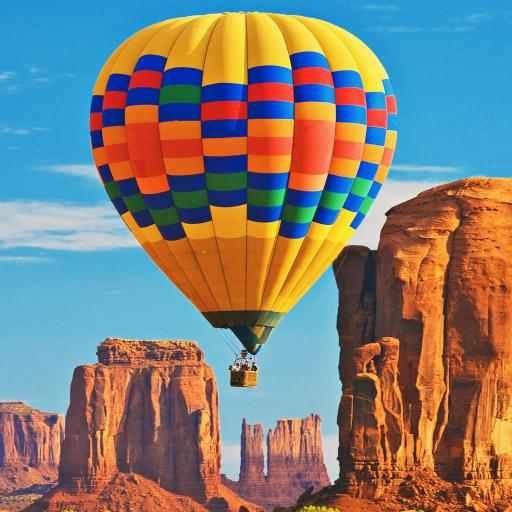}&
		\includegraphics[width=0.115\textwidth]{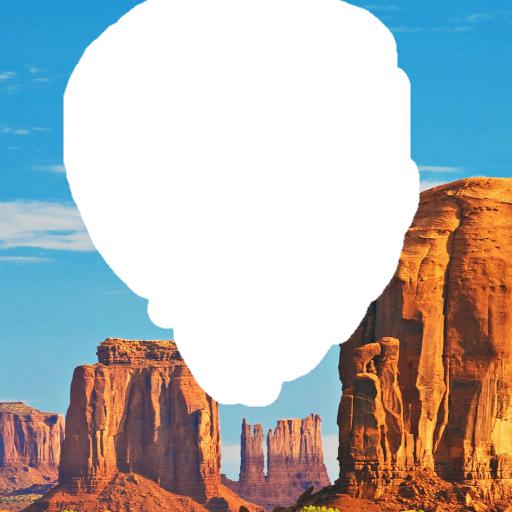}&
		\includegraphics[width=0.115\textwidth]{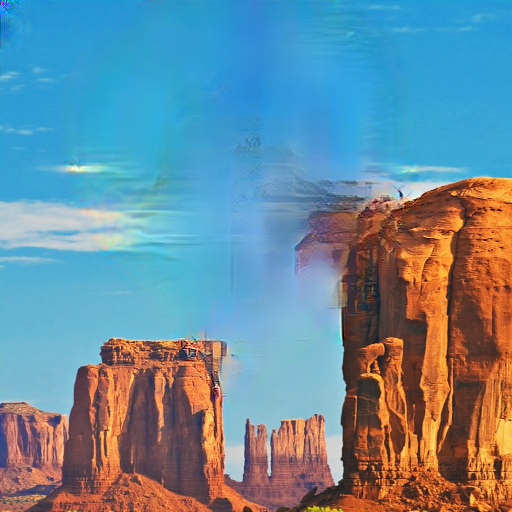}&
		\includegraphics[width=0.115\textwidth]{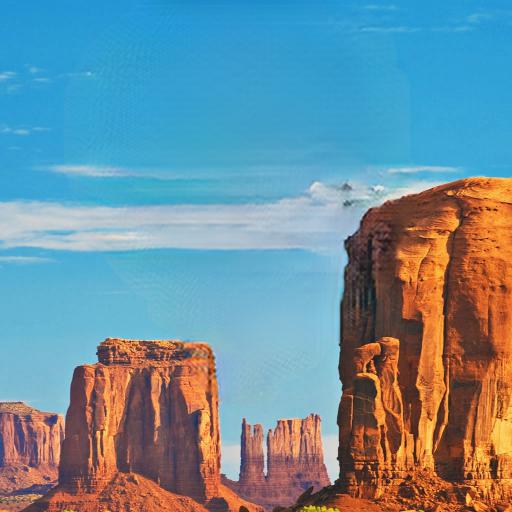}\\
		Original & Input & PConv & Ours \\		
\end{tabular}
\caption{Free-form inpainting results on 512$\times$512 images by PConv and our model. Zoom-in to see the details.}
\label{teaser}
\end{figure}

Existing inpainting methods generally fall into two categories: copying from low-level image features and generating new contents from high-level semantic features. The former ones attempt to fill holes by matching and copying patches from known regions\cite{PatchMatch} or external database images\cite{Hays2018}. These approaches are effective for easy cases, e.g., uniform textured background inpainting. However, they cannot solve challenge cases where missing regions involve complex textures or nonrepetitive structures, due to lacking of high-level semantic understanding\cite{DeepFillV1}. The latter approaches learn to capture the context representation of known regions and synthesize the missing contents by using deep convolutional neural networks (CNN) in an end-to-end training manner. Context Encoder\cite{Pathak} is the first CNN-based method to solve the image inpainting task, wherein a deep convolution encoder-decoder architecture was proposed and an adversarial loss\cite{GAN} was added to generate visually realistic results. The encoder-decoder architecture was extended by many CNN-based inpainting models to get more reasonable and fine-detailed contents\cite{Iizuka,DeepFillV1,DeepFillV2,Pconv,SC-FEGAN,HiFill}. 

However, the end-to-end training encoder-decoder architecture still remains challenging to fill large holes in high resolution images. As shown in Figure \ref{teaser}, the results of PConv\cite{Pconv} are obtained from its online demo\footnote{https://www.nvidia.com/research/inpainting}, which illustrates that PConv fails to generate reasonable contents for these two cases. Because the guidance for filling the holes has lost as some regions in holes are far away from the surrounding known regions, which causes the generator to produce semantically ambiguous contents or visually artifacts for the holes\cite{Li2020}. In addition, higher resolution images make the discriminator easier to tell the fully inpainted images apart from training images, thus drastically amplifying the gradient problem\cite{PGAN}. An alternative way is to use two encoder-decoder architectures to separately fill contents and textures for the unknown regions\cite{DeepFillV1}. Even so, only a few methods can process images with resolution smaller than 512 $\times$ 512, wherein the masked area is generally less than 25\%\cite{DeepFillV2}.

We observe that filling the contents and textures demand quite different abilities from the generator. Content completion relies more on capturing the high-level semantics and the global structure of the image, yet low-level features and local texture statistics of the image is more critical for the texture synthesis. Furthermore, content completion can be regarded as an image generation task\cite{GAN}, and texture synthesis can be treated as an image to image translation task\cite{Pix2pix}. 

Motivated by the observation that high-resolution image inpainting could be disentangled into content completion and texture synthesis, we propose a simple yet powerful high-resolution image inpainting framework named PyramidFill, which attempts to fill large holes in images with high resolution reaching to 1024$\times$1024. Our key insight is that we can  fill the contents for the easier low-resolution images and synthesize the textures for the higher-resolution details progressively, wherein the high-resolution image is formed into a pyramid of different scale images by down-sampling. PyramidFill consists of a pyramid of fully convolutional Generative Adversarial Networks (GANs), wherein the content GAN is responsible for generating contents in the lowest-resolution masked image in image pyramid, and each texture GAN is responsible for synthesizing textures in a higher-resolution image.

Our major contributions can be summarized as follows:
\begin{itemize}
\item We provide a new perspective that high-resolution image inpainting could be disentangled into low-resolution content completion and higher-resolution texture synthesis.
\item Following the new perspective, we design a novel framework consisting of a pyramid of GANs, wherein the content GAN is responsible for generating contents in the lowest-resolution masked image in image pyramid, and each texture GAN is responsible for synthesizing textures in a higher-resolution image.
\item We introduce a new dataset of natural scenery with high resolution 1920 $\times$ 1080 for real image inpainting applicatons.
\end{itemize}
%------------------------------------------------------------------------
\section{Related Work}
\subsection{Deep Image Inpainting}
A variety of CNN-based approaches have been proposed for image inpainting. Pathak et al.\cite{Pathak} first introduced an eoncoder-decoder architecture for the image inpainting task, as well as a pixel-wise reconstruction loss and an adversarial loss. Based on Pathak's work, Iizuka et al.\cite{Iizuka} proposed a fully convolutional GAN model with an extra local discriminator to ensure local image coherency. Yang et al.\cite{Yang2017} proposed a multi-scale neural patch synthesis approach based on joint optimization of image content and texture constraints. Based on two-stage encoder-decoder architectures, Observing ineffectiveness of CNN in modeling long-term correlations between distant contextual information and the hole regions, Yu et al.\cite{DeepFillV1} presented a novel contextual attention layer to integrate in the second stage, which used the features of known patches as convolutional filters to process the generated patches. The aforementioned approaches were based on vanilla convolutions that treated all input pixels as same valid ones, which is not reasonable for masked holes. To address this limitation, Liu et al.\cite{Pconv} proposed a partial convolutional layer for irregular holes in image inpainting, comprising a masked and re-normalized convolution operation followed by a mask-update step. Partial Convolutions could be viewed as hard-mask convolutional layers, Yu et al.\cite{DeepFillV2} proposed gated convolution to learns a dynamic feature gating mechanism for each channel and each spatial location across all layers, which could be viewed as a soft-mask convolutional layer. Yang et al.\cite{YangAAAI} proposed a multi-task learning framework to incorporate the image structure knowledge to assist image inpainting, which trained a shared generator to simultaneous complete the masked image and corresponding structures--- edge and gradient. Liu et al.\cite{MEDFE} proposed a mutual encoder-decoder CNN for joint recovering structures and textures, which used CNN features from the deep and shallow layers of the encoder to represent structures and textures of an input image, respectively. 
\subsection{High-Resolution Image Inpainting}
Most recently, a few works have been presented for high-resolution image inpainting. Instead of directly filling holes in high-resolution images, Yi et al.\cite{HiFill} proposed a contextual residual aggregation mechanism that could produce high-frequency residuals for missing contents by weighted aggregating residuals from contextual patches, thus only requiring a low-resolution prediction from the network. Zeng et al.\cite{Zeng2020} presented a guided upsampling network to generate high-resolution image inpainting results, by extending the contextual attention module which borrowed high-resolution feature patches in the input image.
\subsection{Pyramid in Image Generation}
Pyramid has been explored widely in the image generation task. Denton et al.\cite{LPGAN} introduced a cascade of CNNs within a Laplacian pyramid framework to generate images in a coarse-to-fine fashion. At each level of the pyramid, a separate generation model was trained using the GAN approach. Shaham et al.\cite{SinGAN} introduced SinGAN to learn from a single natural image, which contained a pyramid of fully convolutional GANs, each was responsible for learning the patch distribution at a different scale of the image. Inspired by classical image pyramid representations, Shocher et al.\cite{SemanticP} proposed a semantic generation pyramid framework to generate diverse image samples, which utilized the continuum of semantic information encapsulated in deep features, ranging from low-level textural information contained in fine features to high-level semantic information contained in deeper features. 
%-------------------------------------------------------------------------
\begin{figure*}[t]
	\centering
	\includegraphics[scale=0.4]{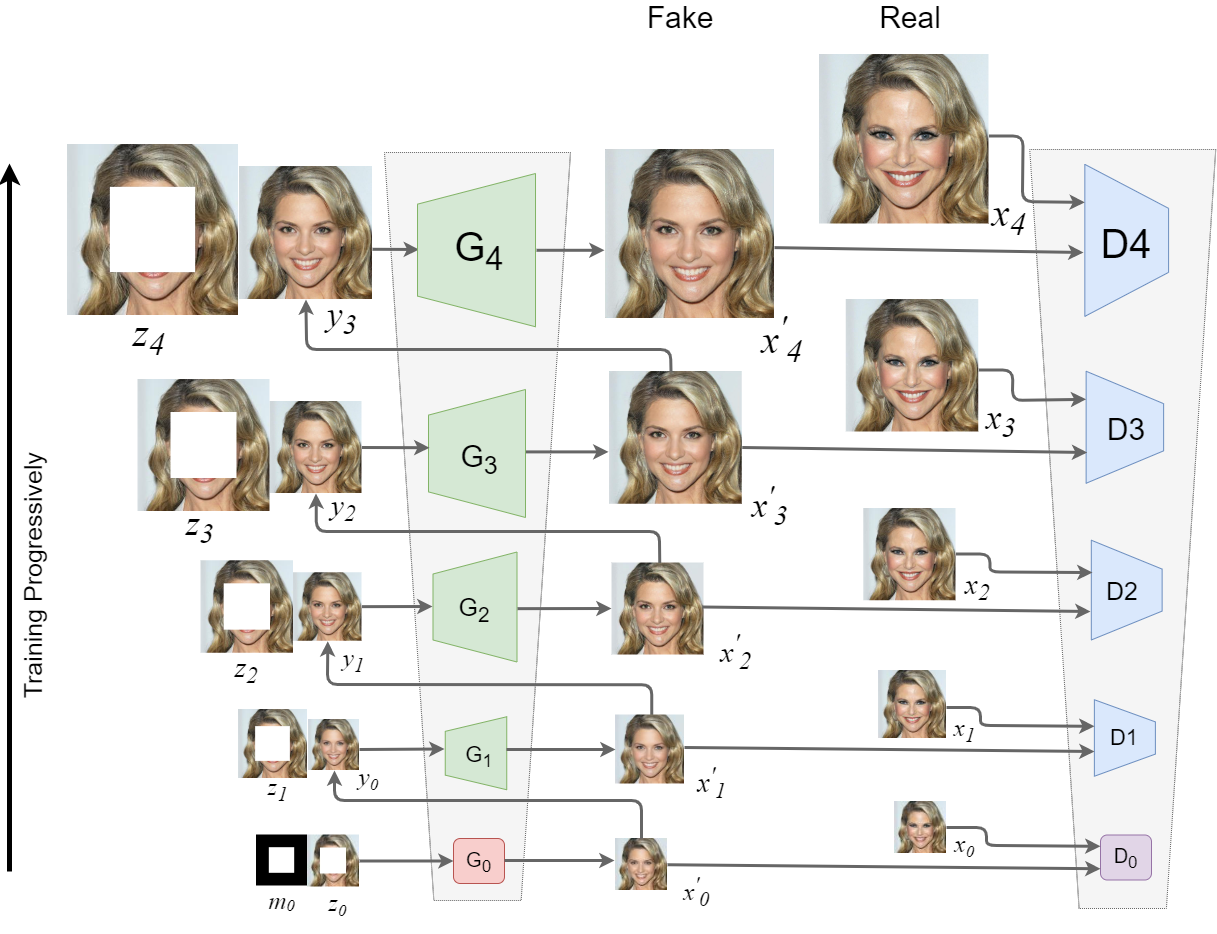}
	\caption{Pipeline of our high-resolution image inpainting algorithm---PyramidFill. Our model consist of a pyramid of PatchGANs, where training are progressive.}
	\label{pipeline}
\end{figure*}
\section{Method}
In this section, we first introduce the pipeline of the proposed PyramidFill, and then present the generator and discriminator networks in each level, as well as the loss functions.
\subsection{Pipeline of PyramidFill}
Figure \ref{pipeline} illustrates the pipeline of our proposed algorithm PyramidFill, which consists of a pyramid of PatchGANs\cite{PatchGAN,Pix2pix}: $\left\{G_{i},D_{i}\right\}$, $i=0,1,2,3,4$. For training, given a high-resolution image $x$, we sample a binary image mask $m$ at a random location (e.g., center area in Figure \ref{pipeline}). Input image $z$ is masked from the original image as $z=x\odot(1-m)+m$. The original image, input image, mask are separately downsampled to be pyramids of images by a factor $r^{4-i}$ (we choose $r=2$): $x_{i}$, $z_{i}$, $m_{i}$ $i=0,1,2,3,4$. Each Generator $G_{i}$ is responsible of filling the corresponding-scale masked image $z_{i}$, trained against the image $x_{i}$, where $G_{i}$ learns to fool the discriminator $D_{i}$.

The filling of a high-resolution image sample starts with the lowest-scale image $x_{0}$, wherein $\left\{G_0,D_0\right\}$ are trained to complete the contents, and then progressively synthesis the finer textures at the higher-scale image by training $\left\{G_i,D_i\right\}$, $i=1,2,3,4$. The generator $G_{0}$ takes concatenation of $z_{0}$ and $m_{0}$ as input, and output the predicted image $x^{'}_{0}$ with the same size as input. We then replace the masked region of $z_{0}$ using the predicted image to get the inpainting result $y_{0}$,
\begin{equation}
\label{eq1}
\begin{split}
y_{0}=z_{0}\odot(1-m_{0})+G_{0}([z_{0},m_{0}])\odot m_{0}
\end{split}
\end{equation}

After training the generator $G_{0}$, the inpainting result $y_{0}$ and input image $z_{1}$ are given to the generator $G_{1}$ to output the predicted image $x^{'}_{1}$ with the same size as $z_{1}$, and it replaces the masked region of $z_{1}$ to get the inpainting result $y_{1}$. Training the other PatchGANs is similar to training $\left\{G_1,D_1\right\}$, given the corresponding input image $z_{i}$ and the lower-scale inpainting result $y_{i-1}$ to get the current-scale inpainting result $y_{i}$, i.e.,
\begin{equation}
\label{eq2}
\begin{split}
y_{i}=z_{i}\odot(1-m_{i})+G_{i}(z_{i},y_{i-1})\odot m_{i}, i=1,2,3,4.
\end{split}
\end{equation}

\begin{figure}[t]
	\centering
	\includegraphics[scale=0.32]{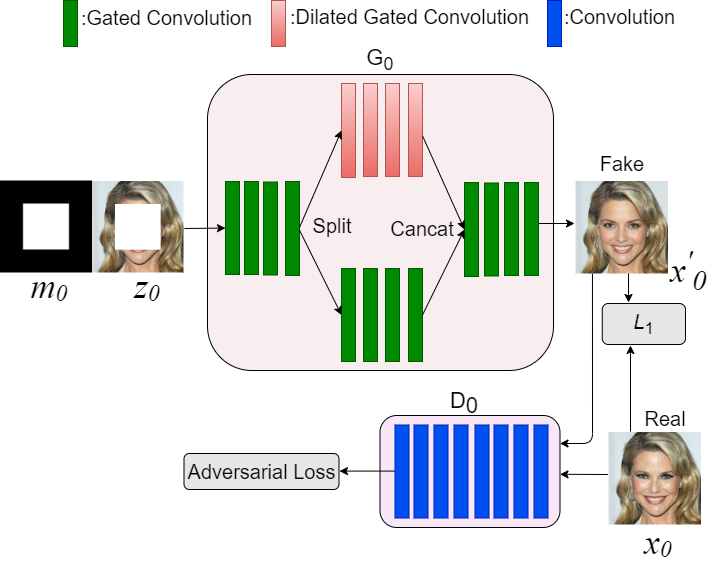}
	\caption{Architecture of a PatchGAN for completing contents.}
	\label{G0}
\end{figure}

\subsection{Generators and Discriminators}
Since completing contents and synthesising textures demand quite different abilities from generators, we customize different architectures for $\left\{G_0,D_0\right\}$ and $\left\{G_i,D_i\right\}$, $i=1,2,3,4$.

Figure \ref{G0} shows the architecture of a PatchGAN for completing contents on the lowest-scale input image $z_{0}$. For the generator network, we maintain the spatial size of the feature maps since of low-scale input images, instead of encoder-decoder networks used in most of image inpainting methods. The input images are first passed through four gated convolutional layers\cite{DeepFillV2}, which are then split into two branches along the channel dimension. Four dilated gated convolutional layers are adopted in the upper branch to expand the size of receptive fields for exploring the global structures of input images, four gated convolutional layers are simultaneously adopted in the lower branch for exploiting fine contents. Feature maps from two branches are then concatenated to proceed the last four gated convolutional layers to predict the contents in the masked region. The discriminator network consists of eight convolutional layers, wherein no downsampling operations are adopted to ensure the spatial size of the output as the same with the input image. It can better capture fine details in the lowest-scale image.

\begin{figure}[t]
	\centering
	\includegraphics[scale=0.32]{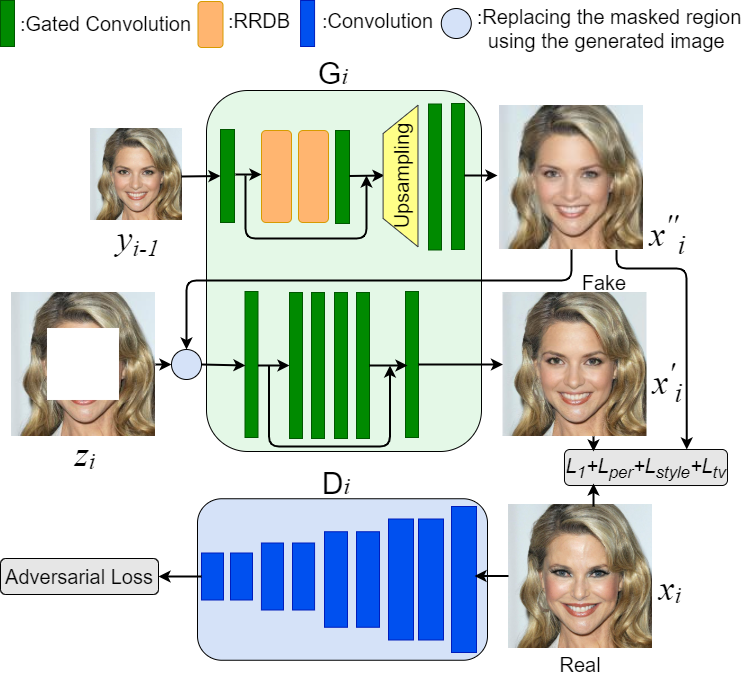}
	\caption{Architecture of a PatchGAN for synthesising textures.}
	\label{G1}
\end{figure}

The architecture of a PatchGAN for synthesising textures is presented in Figure \ref{G1}, and $\left\{G_1,D_1\right\}$, $\left\{G_2,D_2\right\}$, $\left\{G_3,D_3\right\}$, $\left\{G_4,D_4\right\}$ share this same architecture yet progressively training. There are two stages in the generator, and we call the first stage as the super-resolution network, the second stage as the refinement network. The super-resolution network predicts a higher-resolution image from its lower-resolution counterpart which is taken from the inpainting result of the lower-scale generator. We use the predicted higher-resolution image to replace the masked region of the corresponding-scale input image, which then passes through the refinement network to output a finer result. Inspired by ESRGAN\cite{wang2018esrgan} and SRResNet\cite{SRGAN}, we design a simple yet powerful super-resolution network, wherein all vanilla convolutions in RRDB module\cite{wang2018esrgan} are replaced with gated convolutions, and we only adopt two RRDB modules. The upsampling operator uses the sub-pixel layer with the factor 2 to increase the resultion of the input image. In the refinement network, there are only six gated convolutional layers with a shortcut connection within the middle four layers. In the discriminator, we use four strided convolutions with stride 2 to reduce the spatial size of feature maps due to memory cost.

\subsection{Loss Function}
To stabilize the training of PatchGANs, the spectral normalization\cite{SNGAN} is applied in all discriminators. We use a hinge loss\cite{DeepFillV2} as objective function for all PatchGANs $\left\{G_i,D_i\right\}$, $i=0,1,2,3,4$:
\begin{equation}
\label{adv}
\begin{split}
\mathcal{L}^{adv}_{D_{i}} = \mathbb{E}_{x_{i}\sim p_{data}(x_{i})}[\mbox{ReLU}(\mathbbm{1}-D_{i}(x_{i}))] \\
 + \mathbb{E}_{z_{i}\sim p_{z_{i}}(z_{i})}[\mbox{ReLU}(\mathbbm{1}+D_{i}(G_{i}(z_{i})))]
\end{split}
\end{equation}
\begin{equation}
\begin{split}
\mathcal{L}^{adv}_{G_{i}} = -\mathbb{E}_{z_{i}\sim p_{z_{i}}(z_{i})}[D_{i}(G_{i}(z_{i}))]
\end{split}
\end{equation}
where  $x_{i}$ and $z_{i}$ represent the real image and the masked input image, respectively. $D_i$ and $G_i$ denote the spectral-normalized discriminator and the generator, respectively.

Inspired by U-Net GAN\cite{UNET-GAN}, we use a per-pixel consistency regularization technique when training $D_0$, encouraging the discriminator to focus more on semantic and structural changes between real and fake images. This technique can further enhance the quality of the generated sample $x^{'}_{0}$. We train the discriminator to provide consistent per-pixel predictions, by introducing the consistency regularization loss term in the discriminator objective function:
\begin{equation}
\begin{split}
\mathcal{L}^{cons}_{D_{0}} =\Big\|D_{0}(y_{0}) - \Big( D_{0}(x_{0})\odot(1-m_{0}) \\+D_{0}(x^{'}_{0})\odot m_{0}\Big)\Big\|^2
\end{split}
\end{equation}
where $\left\| \cdot \right\|$ denotes the $L^2$ norm. This consistency loss is taken between the per-pixel output of $D_{0}$ on the composing image $y_{0}$ and the composite of output of $D_{0}$ on real and fake images, penalizing the discriminator for inconsistent predictions.

Except the adversarial loss, a pixel reconstruction loss is introduced for training the generators, because the task of generators is not only fool the discriminators but also to generate image being similar to the real image.
\begin{equation}
\label{L1}
\begin{split}
\mathcal{L}^{re}_{x_{gen}} = \frac{1}{N_{x_{gen}}}\left\|(x_{gen}-x_{i})\right\|_{1} 
\end{split}
\end{equation}
where $N_{x_{gen}}$ is the number elements in the image ${x_{gen}}$, which is the predicted image by the generators.

We introduce additional losses for training the texture generators, perceptual loss\cite{Johnson2016Perceptual} and style loss\cite{Pconv,SC-FEGAN}. The perceptual loss penalizes results that are not perceptually similar by computing $L^{1}$ distance between feature maps of a pretrained network,
\begin{equation}
\label{Per}
\begin{split}
\mathcal{L}^{per}_{x_{gen}} = \sum_{q}\frac{1}{N_{q}} \left\| \phi_{q}(x_{gen})-\phi_{q}(x_{i}) \right\|_{1} 
\end{split}
\end{equation}
where $N_{q}$ indicates the number of elements in the $q$-th layer, and $\phi_{q}$ is the feature map of the $q$-th layer extracted from the VGG-16 network pretrained on ImageNet\cite{Simonyan15}, and we choose the feature maps from layers $pool1$, $pool2$ and $pool3$. Style loss compares the content of two images by using Gram matrix. Given feature maps of sizes $C_{q}\times H_{q}\times W_{q}$, style loss is computed as follows:
\begin{equation}
\label{Style}
\begin{split}
\mathcal{L}^{style}_{x_{gen}} = \left\| G^{\phi}_{q}(x_{gen})-G^{\phi}_{q}(x_{i}) \right\|_{1} 
\end{split}
\end{equation}
where $G^{\phi}_{q}$ is a $C_{q}\times C_{q}$ Gram matrix constructed from feature maps $\phi_{q}$, which are the same with that used in the perceptual loss.

The overall loss functions for training $\left\{G_0,D_0\right\}$ are shown as below:
\begin{equation}
\label{D0}
\begin{split}
\mathcal{L}_{D_{0}}=\mathcal{L}^{adv}_{D_{0}}+\mathcal{L}^{cons}_{D_{0}} \\
\mathcal{L}_{G_{0}}=\mathcal{L}^{adv}_{G_{0}}+\mathcal{L}^{re}_{x^{'}_{0}}
\end{split}
\end{equation}

And the whole loss functions for training $\left\{G_i,D_i\right\}, i=1,2,3,4$ are as follows:
\begin{equation}
\label{Di}
\begin{split}
\mathcal{L}_{D_{i}}=\mathcal{L}^{adv}_{D_{i}}
\end{split}
\end{equation}
\begin{equation}
\label{Gi}
\begin{split}
\mathcal{L}_{G_{i}}=\lambda_{a}\mathcal{L}^{adv}_{G_{i}}+\lambda_{r}\mathcal{L}^{re}_{x^{''}_{i}}+\lambda_{p}\mathcal{L}^{per}_{x^{''}_{i}}+\lambda_{s_{i}}\mathcal{L}^{style}_{x^{''}_{i}} \\
+\lambda_{r}\mathcal{L}^{re}_{x^{'}_{i}}+\lambda_{p}\mathcal{L}^{per}_{x^{'}_{i}}+\lambda_{s_{i}}\mathcal{L}^{style}_{x^{'}_{i}}
\end{split}
\end{equation}
For our experiments, we choose $\lambda_{a}=0.001$, $\lambda_{r}=0.1$, $\lambda_{p}=0.1$ for all generators, and $\lambda_{s_{1}}=1$, $\lambda_{s_{2}}=50$, $\lambda_{s_{3}}=120$, $\lambda_{s_{4}}=250$ for different generators, respectively.
\section{Experiments}
We evaluate PyramidFill on three datasets: CelebA-HQ\cite{PGAN}, Places2\cite{zhou2017places}, and  our new collected NSHQ dataset. CelebA-HQ contains 30,000 high-quality images at 1024$\times$1024 resolution focusing on human faces. For Places2, we randomly select 20 categories from the 365 categories to form a subset of 100,000 images. The NSHQ dataset includes 5000 high-quality natural scenery images at 1920$\times$1080 or 1920$\times$1280 resolution. For Places2 and NSHQ, images are randomly cropped to 512$\times$512 and 1024$\times$1024, respectively.  Therefore, there is no $\left\{G_4,D_4\right\}$ for training Places2 subset. For all datasets, we randomly partition into 90\% of images for training and 10\% of images for testing. 

We compare PyramidFill with five state-of-the-art approaches: Global\&Local\cite{Iizuka}, DeepFillV1\cite{DeepFillV1}, PConv\cite{Pconv}, PEN-Net\cite{Zeng2019}, DeepFillV2\cite{DeepFillV2}. When training CelebA-HQ, we use regular masks and random free-form masks\cite{DeepFillV2}, wherein the regular masks cover image center with half of image size. For Places2 and NSHQ, random free-form masks\cite{DeepFillV2} are employed for training, while irregular masks from PConv\cite{Pconv} are used for testing.

\begin{table}[ht]
	\centering
	\caption{Quantitative comparisons on CelebA-HQ testing set with input size 256$\times$256 resolution, where the inputs are with center hole regions. The $\downarrow$ indicates lower is better while $\uparrow$ indicates higher is better.}
	\label{Quantitative1}
	\vspace{0.1in}
	\begin{tabular}
		{c|c|c|c}
		\hline
		      Methods  & L1 Loss$\downarrow$ & PSNR$\uparrow$ & SSIM$\uparrow$  \\
		\hline
		 Global\&Local & 0.0386  & 24.09 & 0.8328 \\
		 DeepFillV1    & 0.0418  & 23.75 & 0.8348 \\
		 PConv         & 0.0262  & 25.18 & 0.8757 \\
		 PEN-Net       & 0.0322  & 25.06 & 0.8536 \\
		 DeepFillV2    & 0.0367  & 25.19 & 0.8527  \\
 		 Ours          & \textbf{0.0227}  & \textbf{26.01} & \textbf{0.8878}  \\
		\hline
	\end{tabular}
	%\vspace{-0.1in}
\end{table}

\begin{table}[ht]
	\centering
	\caption{Quantitative comparisons on CelebA-HQ dataset with input size 512$\times$512 resolution, where the inputs are with center hole regions.}
	\label{Quantitative2}
	\vspace{0.1in}
	\begin{tabular}
		{c|c|c|c}
		\hline
		      Methods  & L1 Loss$\downarrow$ & PSNR$\uparrow$ & SSIM$\uparrow$  \\
		\hline
		 PConv         & 0.0302  & 23.07 & 0.8791 \\
		 DeepFillV2    & 0.0344  & 22.23 & 0.8707  \\
 		 Ours          & \textbf{0.0237}  & \textbf{25.61} & \textbf{0.8860}  \\
		\hline
	\end{tabular}
	%\vspace{-0.1in}
\end{table}

\begin{table}[ht]
	\centering
	\caption{Quantitative comparisons on Places2 subset with input size 256$\times$256 resolution, where the inputs are with irregular holes.}
	\label{Quantitative3}
	\vspace{0.1in}
	\begin{tabular}
		{c|c|c|c|c}
		\hline
		   Mask & Methods & L1 Loss$\downarrow$ & PSNR$\uparrow$  & SSIM$\uparrow$  \\
		\hline
		 \multirow{6}{*}{0-10\%} & Global\&Local  &  0.0109 & 29.43 & 0.9443  \\
		 & DeepFillV1     & 0.0101  & 31.51 & 0.9571 \\
		 & PConv         & 0.0091  & 32.27 & 0.9487 \\
		 & PEN-Net       & 0.0079  & 32.66 & 0.9622 \\
		 & DeepFillV2    & 0.0069  & 35.77 & 0.9727 \\
 		 & Ours          & \textbf{0.0016}  & \textbf{38.30} & \textbf{0.9864}  \\		 
		\hline
		\multirow{6}{*}{10-20\%} & Global\&Local  &  0.0231 & 24.90 & 0.8724  \\
		 & DeepFillV1     & 0.0225  & 25.44 & 0.8928 \\
		 & PConv         & 0.0172  & 28.13 & 0.8927 \\
		 & PEN-Net       & 0.0190  & 26.84 & 0.9017 \\
		 & DeepFillV2    & 0.0136  & 30.22 & 0.9311  \\
 		 & Ours          & \textbf{0.0091}  & \textbf{31.86} & \textbf{0.9162}  \\		 
		\hline
		\multirow{6}{*}{20-30\%} & Global\&Local  &  0.0372 & 22.42 & 0.7963  \\
		 & DeepFillV1     & 0.0384  & 22.25 & 0.8179 \\
		 & PConv         & 0.0265  & 25.68 & 0.8334 \\
		 & PEN-Net       & 0.0331  & 23.71 & 0.8287 \\
		 & DeepFillV2    & 0.0227  & 26.85 & 0.8768  \\
 		 & Ours          & \textbf{0.0129}  & \textbf{29.12} & \textbf{0.8569}  \\		 
		\hline
		\multirow{6}{*}{30-40\%} & Global\&Local  &  0.0514 & 20.73 & 0.7265  \\
		 & DeepFillV1     & 0.0544  & 20.27 & 0.7471 \\
		 & PConv         & 0.0360  & 23.96 & 0.7765 \\
		 & PEN-Net       & 0.0479  & 21.66 & 0.7588 \\
		 & DeepFillV2    & 0.0325  & 24.62 & 0.8213  \\
 		 & Ours          & \textbf{0.0132}  & \textbf{30.06} & \textbf{0.8499}  \\		 
		\hline
		\multirow{6}{*}{40-50\%} & Global\&Local  &  0.0665 & 19.40 & 0.6559  \\
		 & DeepFillV1     & 0.0706  & 18.90 & 0.6757 \\
		 & PConv         & 0.0475  & 22.40 & 0.7121 \\
		 & PEN-Net       & 0.0656  & 19.94 & 0.6839 \\
		 & DeepFillV2    & 0.0441  & 22.79 & 0.7596 \\
 		 & Ours          & \textbf{0.0246}  & \textbf{26.19} & \textbf{0.7680}  \\		 
		\hline
	\end{tabular}
	%\vspace{-0.1in}
\end{table}

\begin{figure*}[ht]
	\centering
	\begin{tabular}
		{@{\hspace{0.0mm}}c@{\hspace{0.7mm}}c@{\hspace{0.7mm}}c@{\hspace{0.7mm}}c@{\hspace{0.7mm}}c@{\hspace{0.7mm}}c@{\hspace{0.7mm}}c@{\hspace{0.7mm}}c@{\hspace{0mm}}}
		\includegraphics[width=0.12\linewidth]{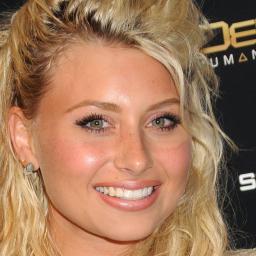}&
		\includegraphics[width=0.12\linewidth]{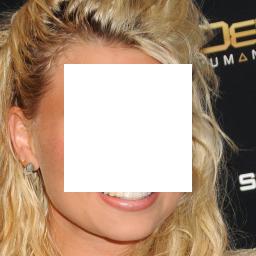}&
		\includegraphics[width=0.12\linewidth]{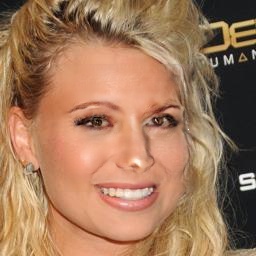}&
		\includegraphics[width=0.12\linewidth]{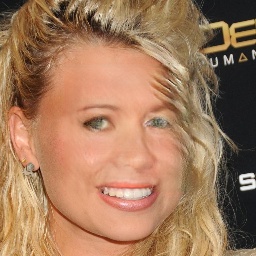}&
		\includegraphics[width=0.12\linewidth]{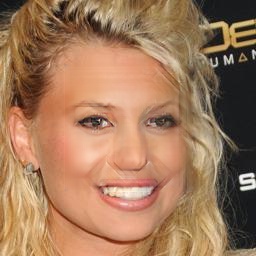}&
		\includegraphics[width=0.12\linewidth]{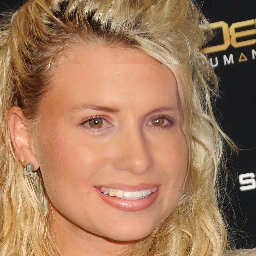}&
		\includegraphics[width=0.12\linewidth]{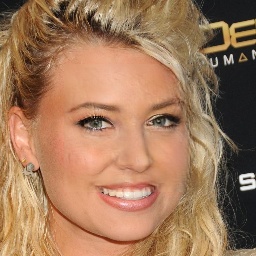}&
		\includegraphics[width=0.12\linewidth]{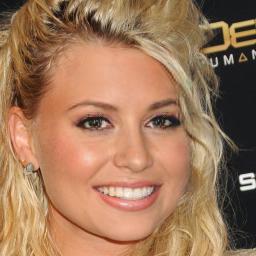} \\
		\includegraphics[width=0.12\linewidth]{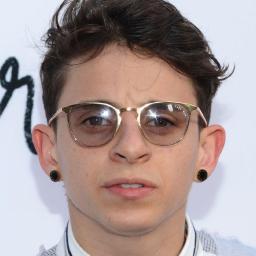}&
		\includegraphics[width=0.12\linewidth]{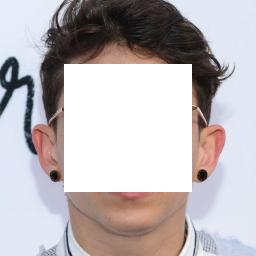}&
		\includegraphics[width=0.12\linewidth]{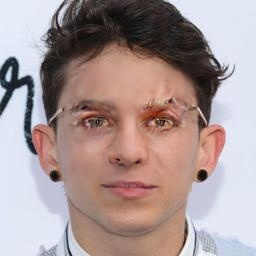}&
		\includegraphics[width=0.12\linewidth]{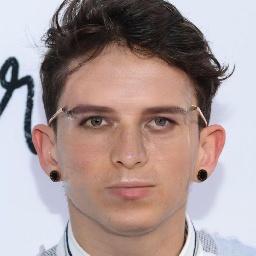}&
		\includegraphics[width=0.12\linewidth]{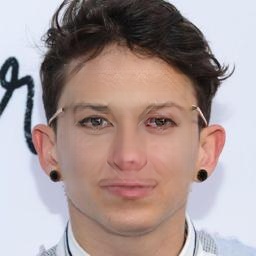}&
		\includegraphics[width=0.12\linewidth]{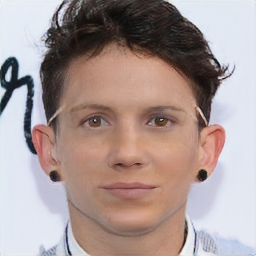}&
		\includegraphics[width=0.12\linewidth]{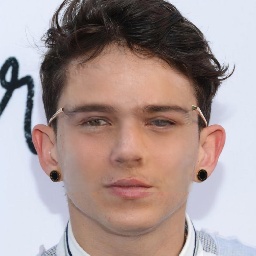}&
		\includegraphics[width=0.12\linewidth]{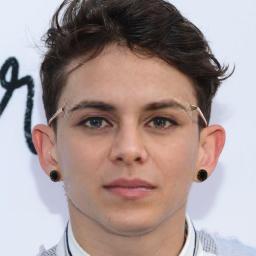} \\
		\includegraphics[width=0.12\linewidth]{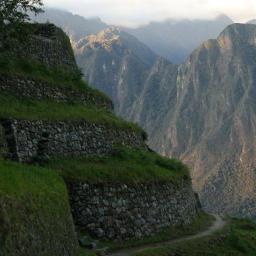}&
		\includegraphics[width=0.12\linewidth]{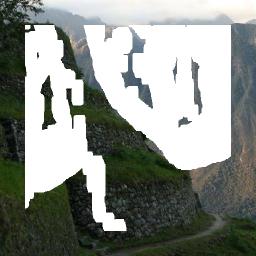}&
		\includegraphics[width=0.12\linewidth]{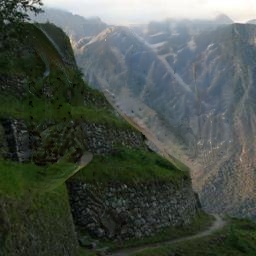}&
		\includegraphics[width=0.12\linewidth]{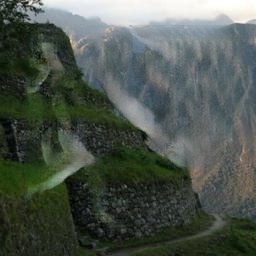}&
		\includegraphics[width=0.12\linewidth]{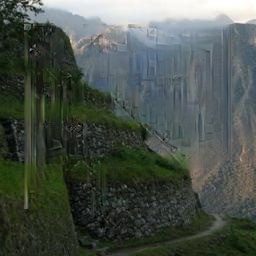}&
		\includegraphics[width=0.12\linewidth]{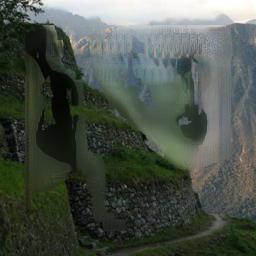}&
		\includegraphics[width=0.12\linewidth]{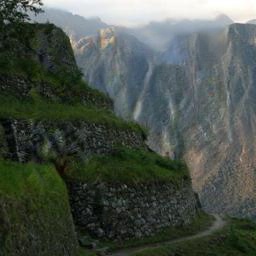}&
		\includegraphics[width=0.12\linewidth]{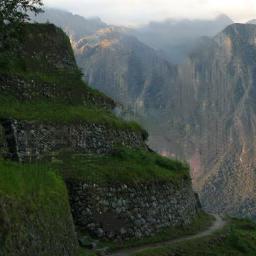} \\
		\includegraphics[width=0.12\linewidth]{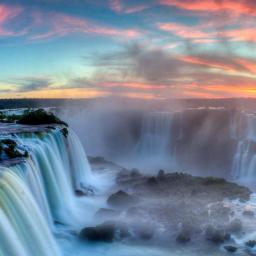}&
		\includegraphics[width=0.12\linewidth]{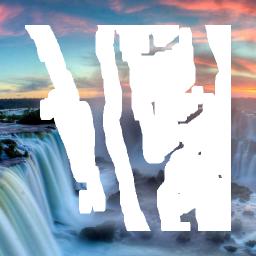}&
		\includegraphics[width=0.12\linewidth]{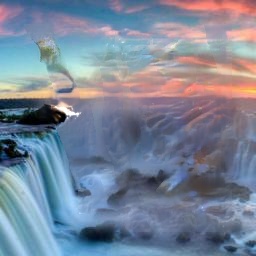}&
		\includegraphics[width=0.12\linewidth]{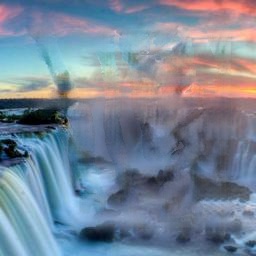}&
		\includegraphics[width=0.12\linewidth]{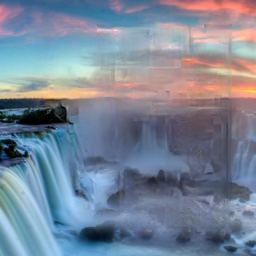}&
		\includegraphics[width=0.12\linewidth]{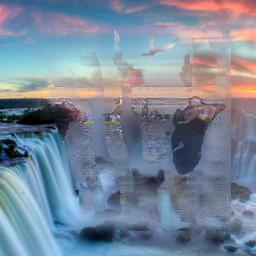}&
		\includegraphics[width=0.12\linewidth]{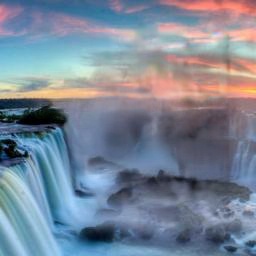}&
		\includegraphics[width=0.12\linewidth]{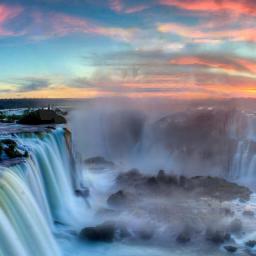} \\
		\includegraphics[width=0.12\linewidth]{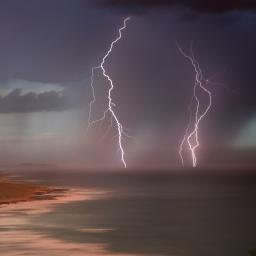}&
		\includegraphics[width=0.12\linewidth]{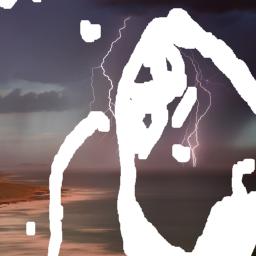}&
		\includegraphics[width=0.12\linewidth]{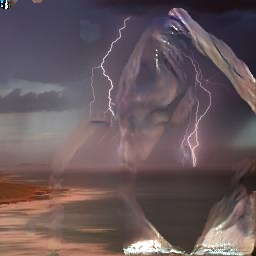}&
		\includegraphics[width=0.12\linewidth]{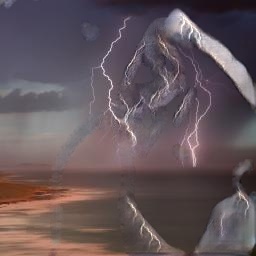}&
		\includegraphics[width=0.12\linewidth]{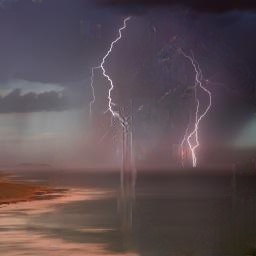}&
		\includegraphics[width=0.12\linewidth]{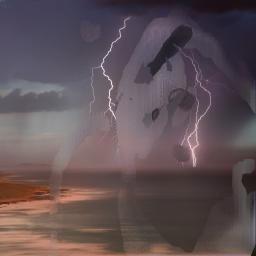}&
		\includegraphics[width=0.12\linewidth]{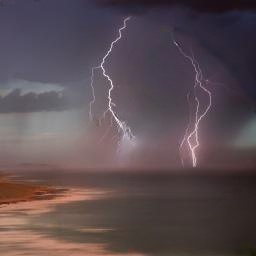}&
		\includegraphics[width=0.12\linewidth]{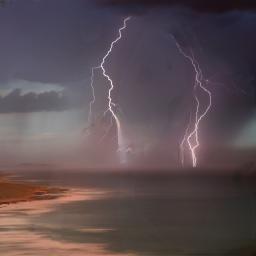} \\
		\includegraphics[width=0.12\linewidth]{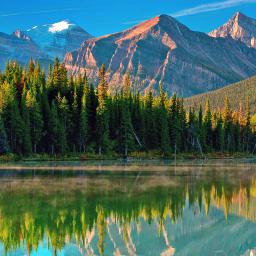}&
		\includegraphics[width=0.12\linewidth]{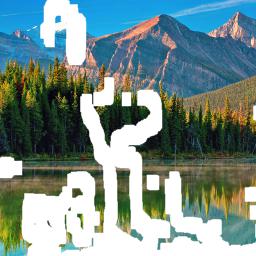}&
		\includegraphics[width=0.12\linewidth]{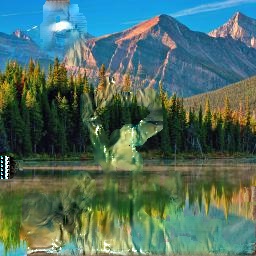}&
		\includegraphics[width=0.12\linewidth]{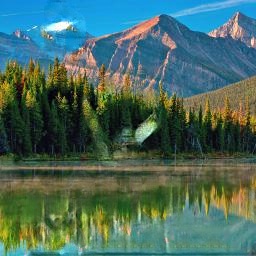}&
		\includegraphics[width=0.12\linewidth]{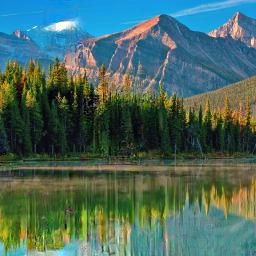}&
		\includegraphics[width=0.12\linewidth]{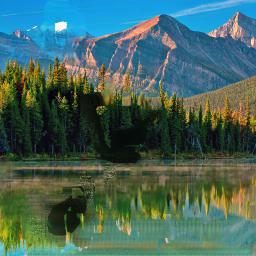}&
		\includegraphics[width=0.12\linewidth]{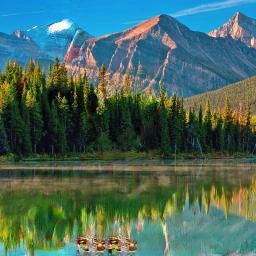}&
		\includegraphics[width=0.12\linewidth]{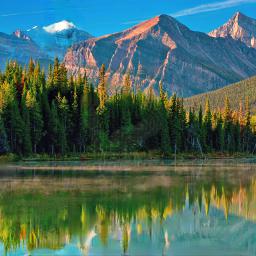} \\
		Original & Input & Global\&Local & DeepFillV1 & PConv & PEN-Net & DeepFillV2 & Ours \\
				
	\end{tabular}
	\vspace{0.1in}
	\caption{Example cases of qualitative comparisons on CelebA-HQ, Places2 and NSHQ testing sets with 256$\times$256 images.}
	\label{Qualitative}
\end{figure*}

\begin{table}[ht]
	\centering
	\caption{Quantitative comparisons on Places2 subset with input size 512$\times$512 resolution, where the inputs are with irregular holes.}
	\label{Quantitative4}
	\vspace{0.1in}
	\begin{tabular}
		{c|c|c|c|c}
		\hline
		   Mask & Methods & L1 Loss$\downarrow$ & PSNR$\uparrow$  & SSIM$\uparrow$  \\
		\hline
		 \multirow{3}{*}{0-10\%} & PConv & 0.0068 & 34.05 & 0.9647 \\
		 & DeepFillV2    & 0.0070  & 34.82 & 0.9710 \\
 		 & Ours          & \textbf{0.0012}  & \textbf{37.93} & \textbf{0.9869}  \\		 
		\hline
		\multirow{3}{*}{10-20\%} & PConv  & 0.0141  & 29.11 & 0.9168 \\
		 & DeepFillV2    & 0.0146  & 29.14 & 0.9272  \\
 		 & Ours          & \textbf{0.0075}  & \textbf{31.19} & \textbf{0.9263}  \\		 
		\hline
		\multirow{3}{*}{20-30\%} & PConv  & 0.0238  & 26.09 & 0.8584 \\
		 & DeepFillV2    & 0.0248  & 25.80 & 0.8716  \\
 		 & Ours          & \textbf{0.0132}  & \textbf{28.09} & \textbf{0.8460}  \\		 
		\hline
		\multirow{3}{*}{30-40\%}& PConv  & 0.0343  & 23.96 & 0.8006 \\
		 & DeepFillV2    & 0.0343  & 23.59 & 0.8156  \\
 		 & Ours          & \textbf{0.0131}  & \textbf{28.92} & \textbf{0.8433}  \\		 
		\hline
		\multirow{3}{*}{40-50\%} & PConv  & 0.0472  & 22.09 & 0.7360 \\
		 & DeepFillV2    & 0.0487  & 21.79 & 0.7548 \\
 		 & Ours          & \textbf{0.0241}  & \textbf{25.47} & \textbf{0.7549}  \\		 
		\hline
	\end{tabular}
	%\vspace{-0.1in}
\end{table}

\subsection{Quantitative Evaluation}
We conduct quantitative comparisons on CelebA-HQ dataset and Places2 subset. For a fair evaluation, all input images are resized to 256$\times$256 and 512$\times$512, respectively. Because a few compared official pre-trained models are only trained on 256$\times$256 images. Table \ref{Quantitative1} and Table \ref{Quantitative2} report the quantitative comparisons on CelebA-HQ testing set with 256$\times$256 images and 512$\times$512 images, respectively. The images are masked with center holes, and we use L1 loss, PSNR and SSIM as metrics. It shows that our PyramidFill outperforms existing state-of-the-art methods with evident superiorities. Table \ref{Quantitative3} and Table \ref{Quantitative4} present the quantitative comparisons on Places2 testing set with 256$\times$256 images and 512$\times$512 images masked with irregular masks, respectively. The results are categorized according to the ratios of the hole regions versus the image size. It shows that our PyramidFill performs much better than the state-of-the-art methods, especially for filling large holes in images.

\subsection{Qualitative Evaluation}
The qualitative comparisons on three datasets are shown in Figure\ref{Qualitative}. It shows that Global\&Local and DeepFillV1 often generate heavy artifacts, even the holes are not very large. PConv, PEN-Net and DeepFillV2 can generate correct contents for completing faces, yet lacking of detailed textures. By contrast, our model PyramidFill can generate more realistic results on the corrupted faces with finer textures. When dealing with the corrupted images from Places2 and NSHQ, PConv and PEN-Net also produce obvious artifacts. DeepFillV2 and our PyramidFill both can generate reasonable contents and detailed textures, but results from our PyramidFill have better similarity to the original images than results from DeepFillV2.

\begin{figure}[t!]
	\centering
	\begin{tabular}
		{@{\hspace{0.0mm}}c@{\hspace{0.7mm}}c@{\hspace{0.7mm}}c@{\hspace{0.7mm}}c@{\hspace{0mm}}}
		\includegraphics[width=0.32\linewidth]{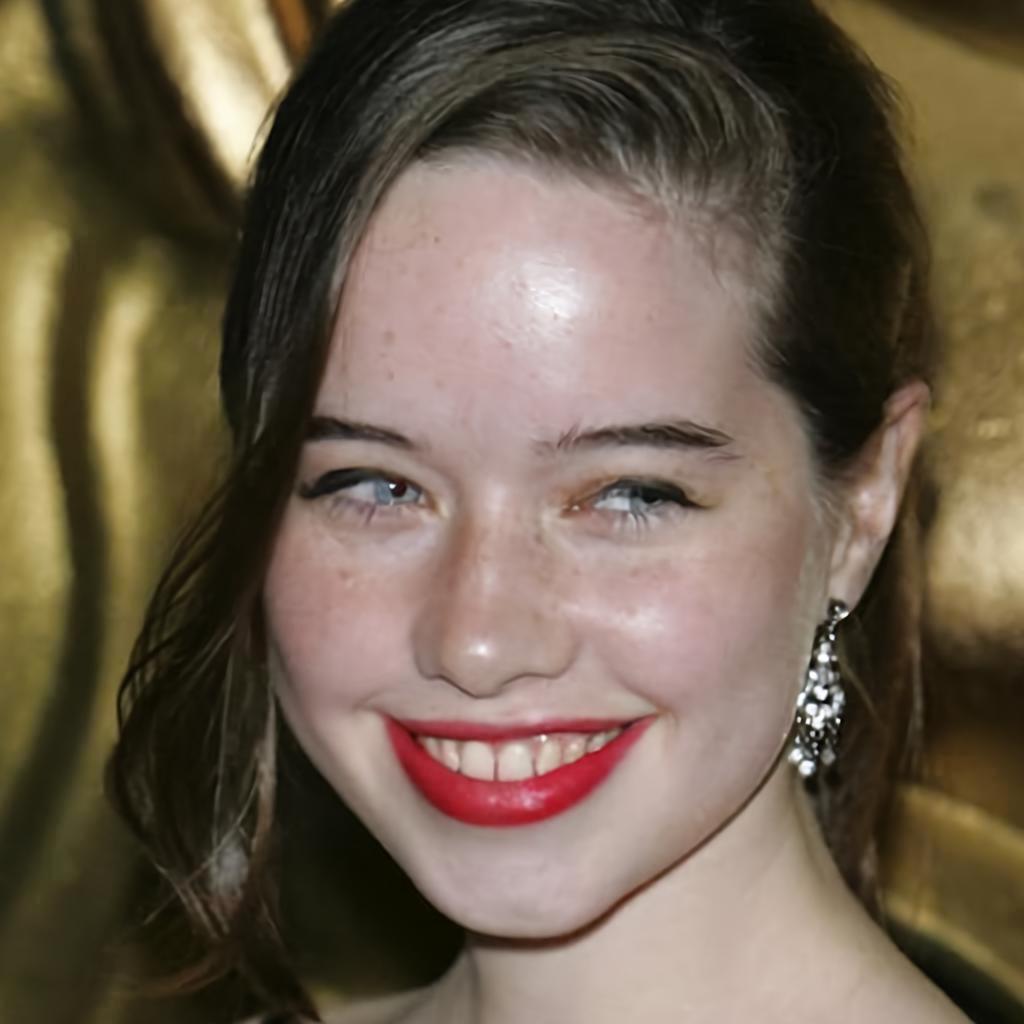}&
		\includegraphics[width=0.32\linewidth]{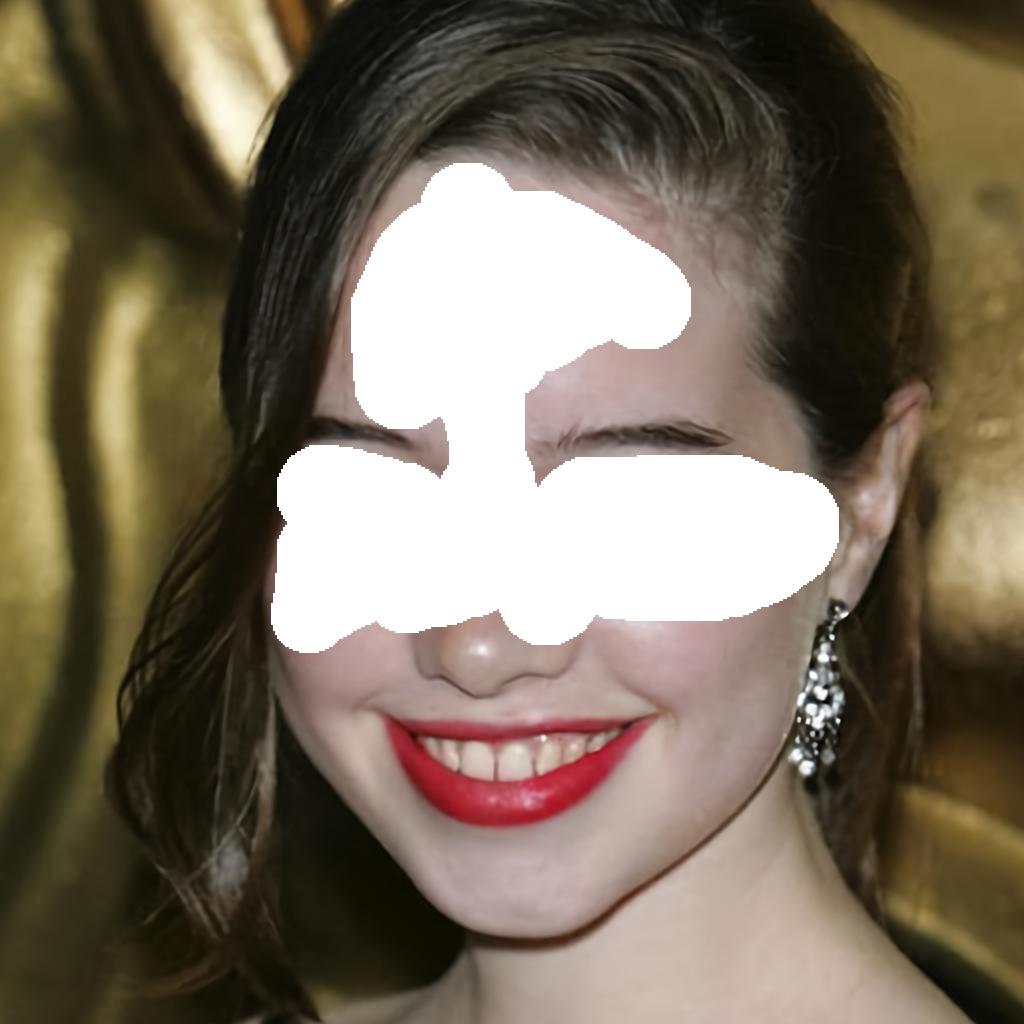}&
		\includegraphics[width=0.32\linewidth]{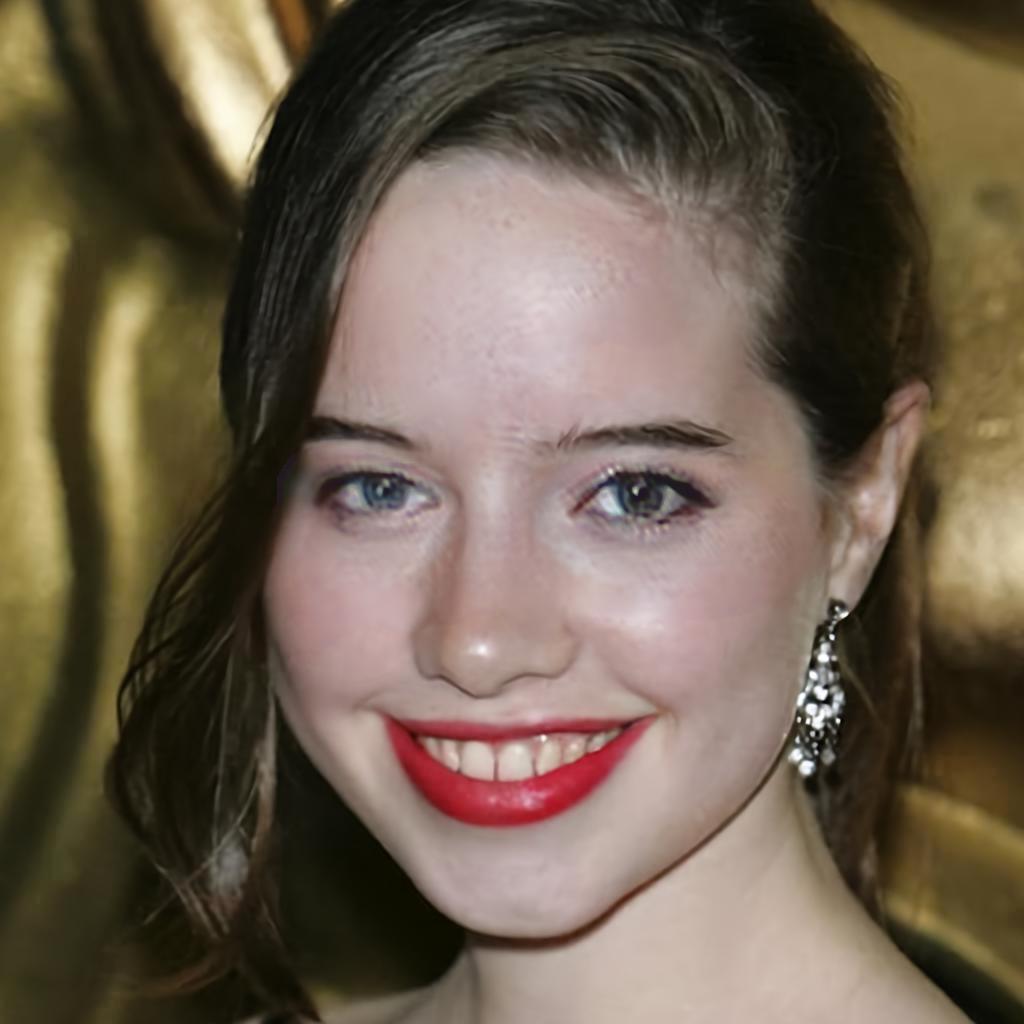} \\
		\includegraphics[width=0.32\linewidth]{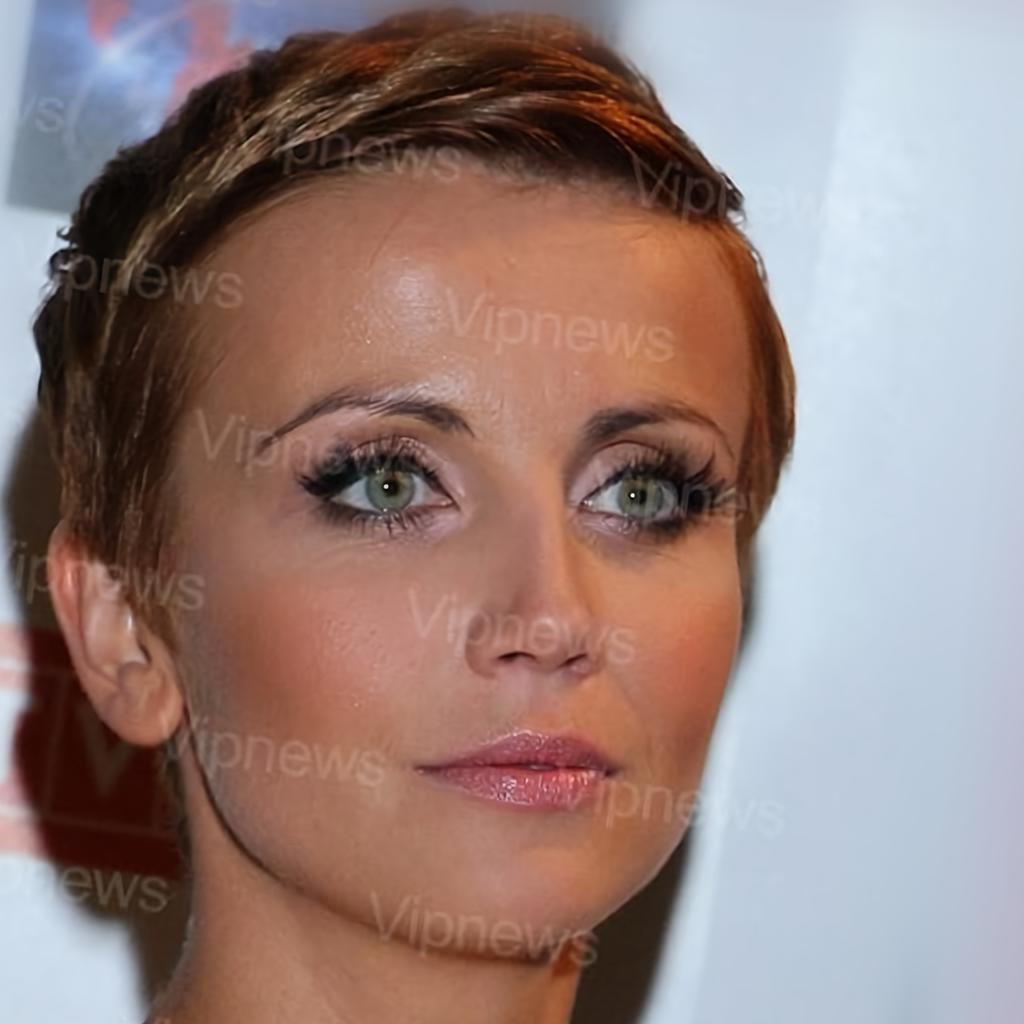}&
		\includegraphics[width=0.32\linewidth]{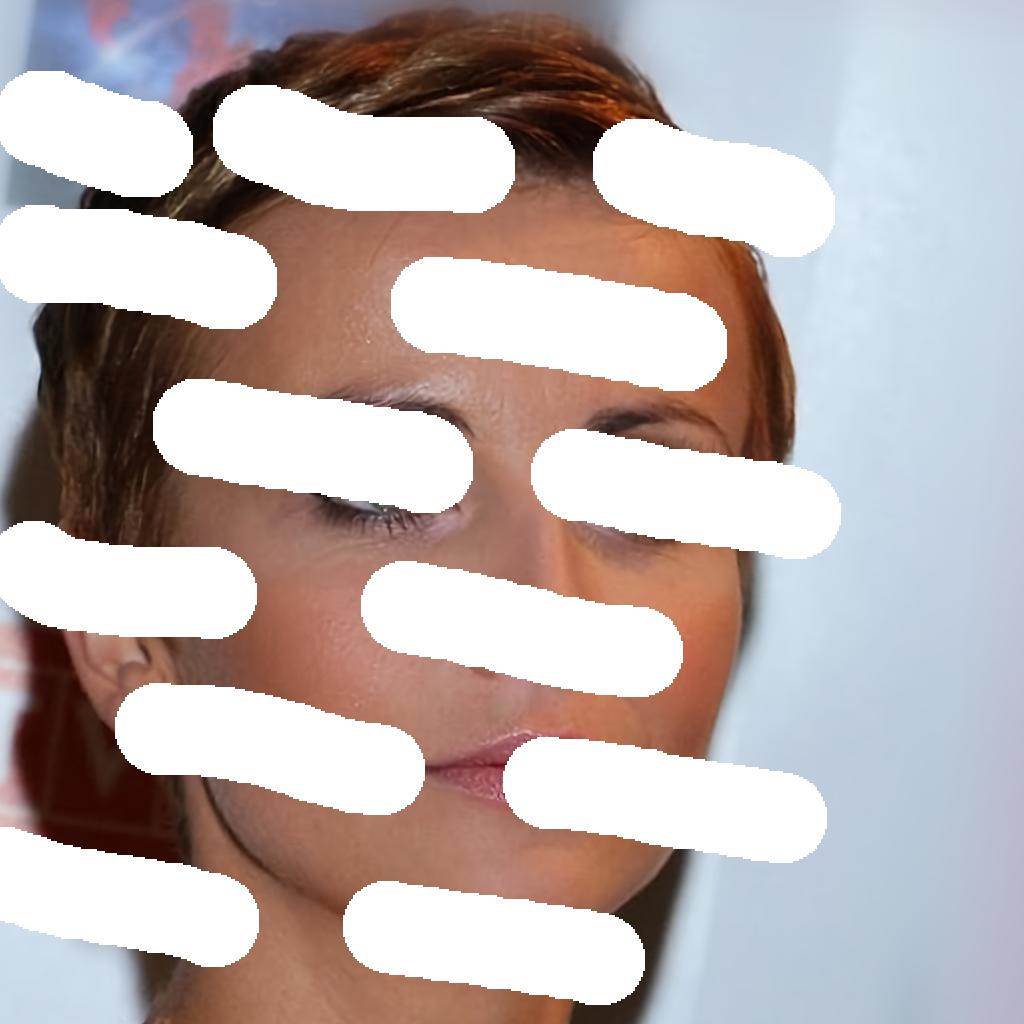}&
		\includegraphics[width=0.32\linewidth]{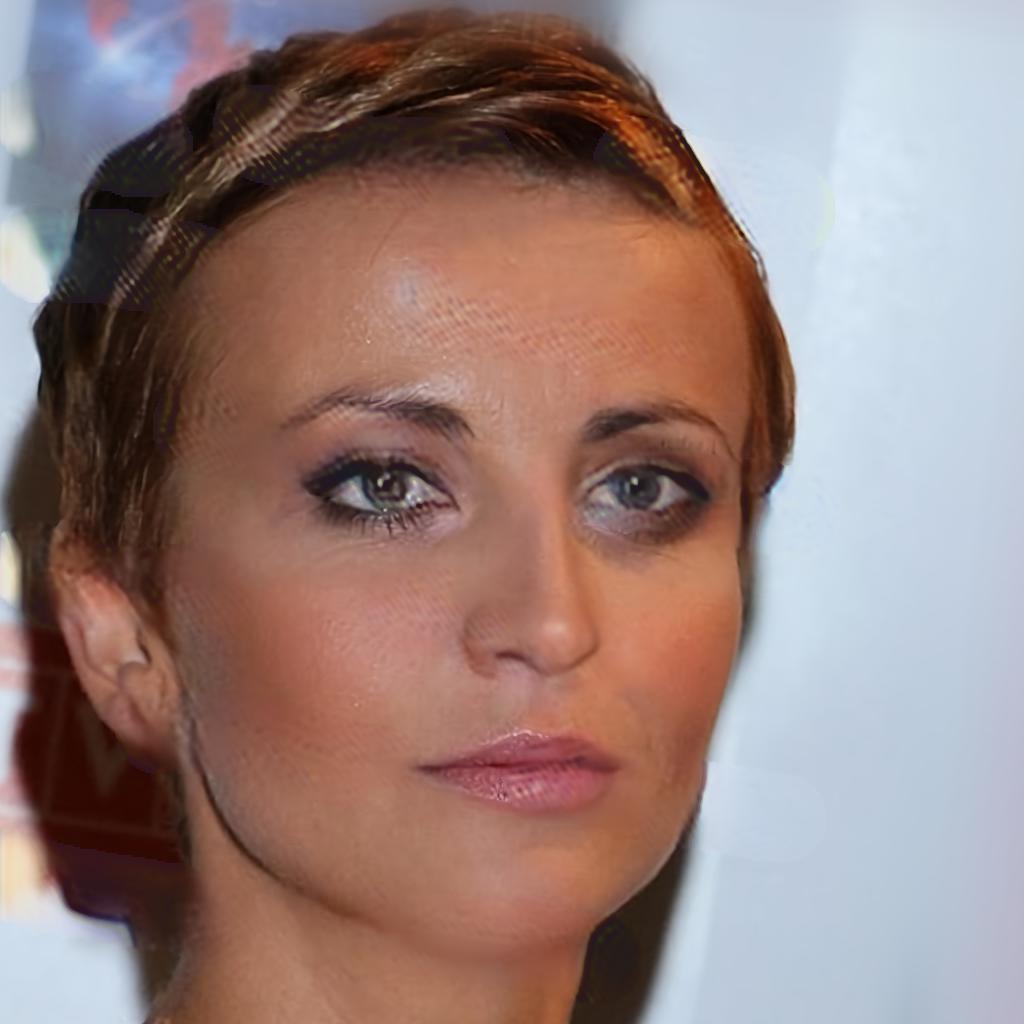} \\
		\includegraphics[width=0.32\linewidth]{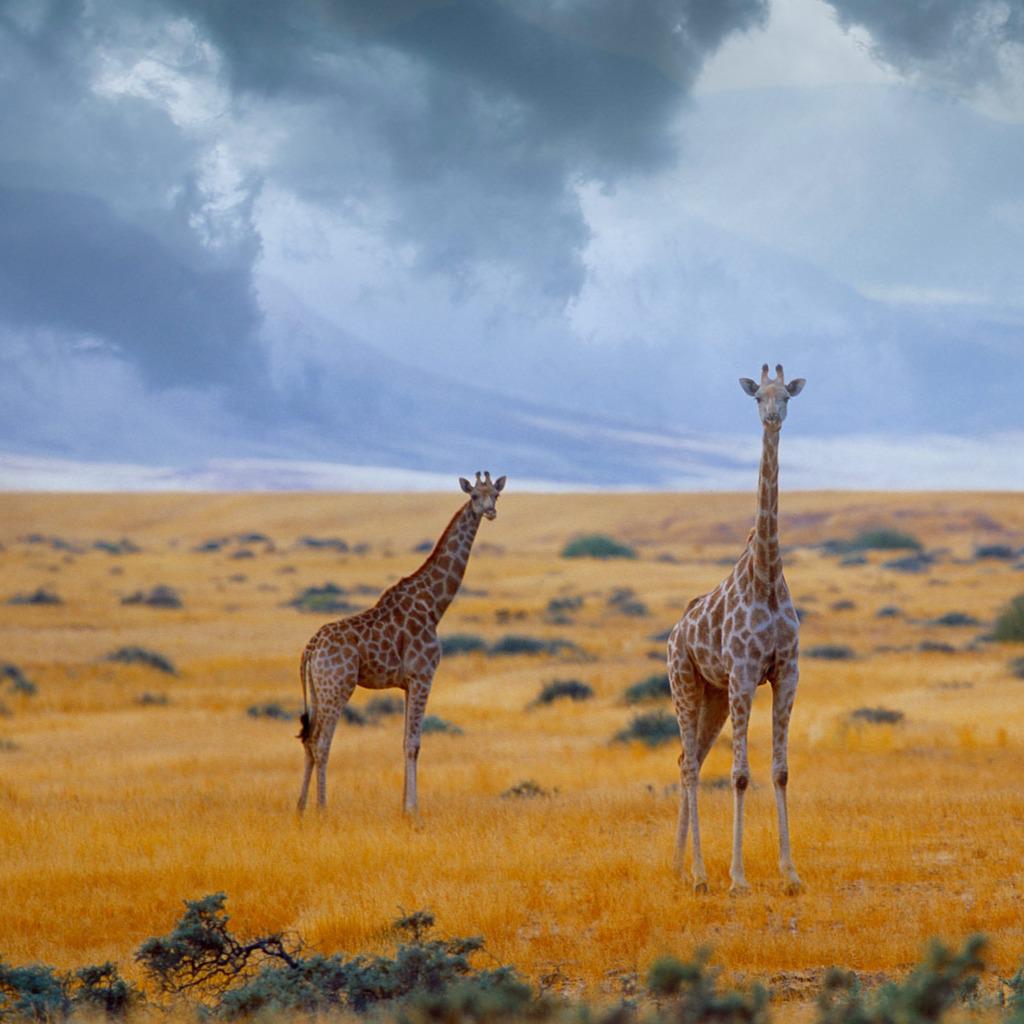}&
		\includegraphics[width=0.32\linewidth]{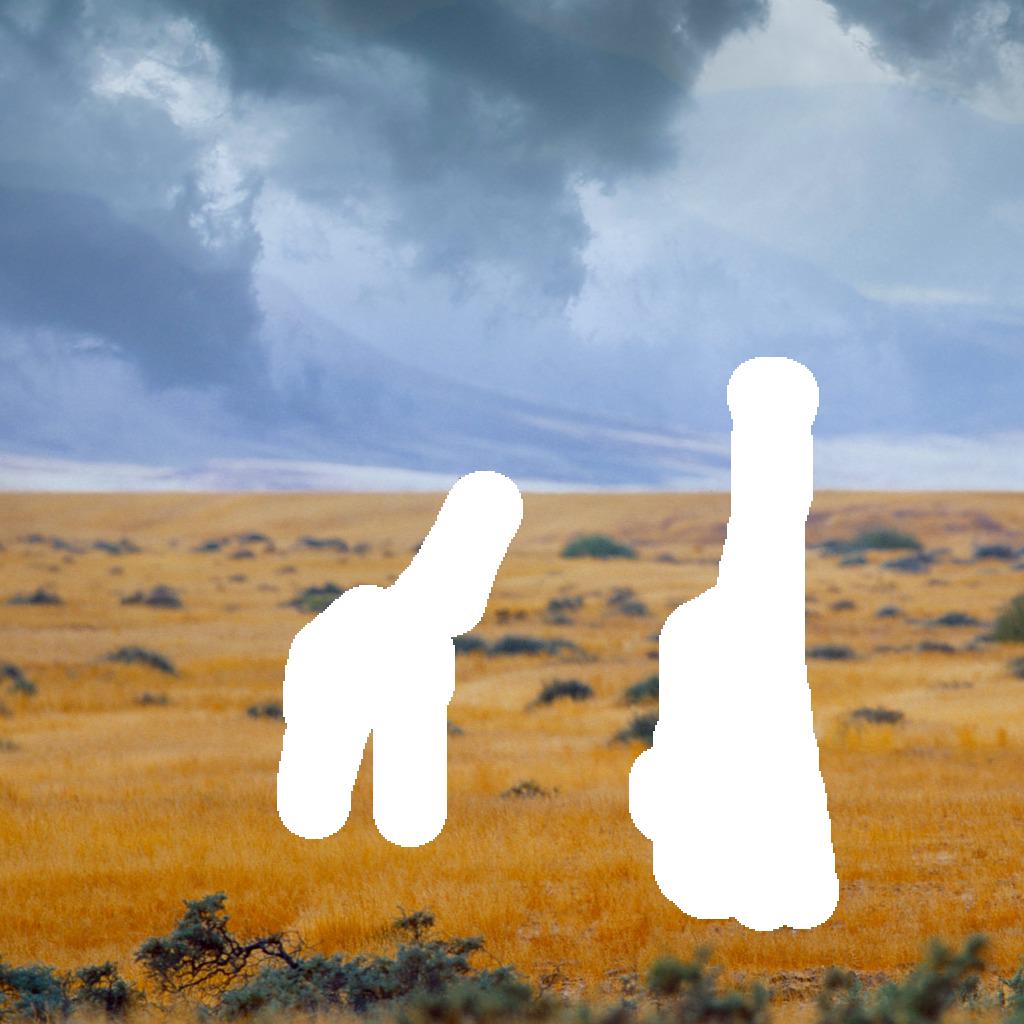}&
		\includegraphics[width=0.32\linewidth]{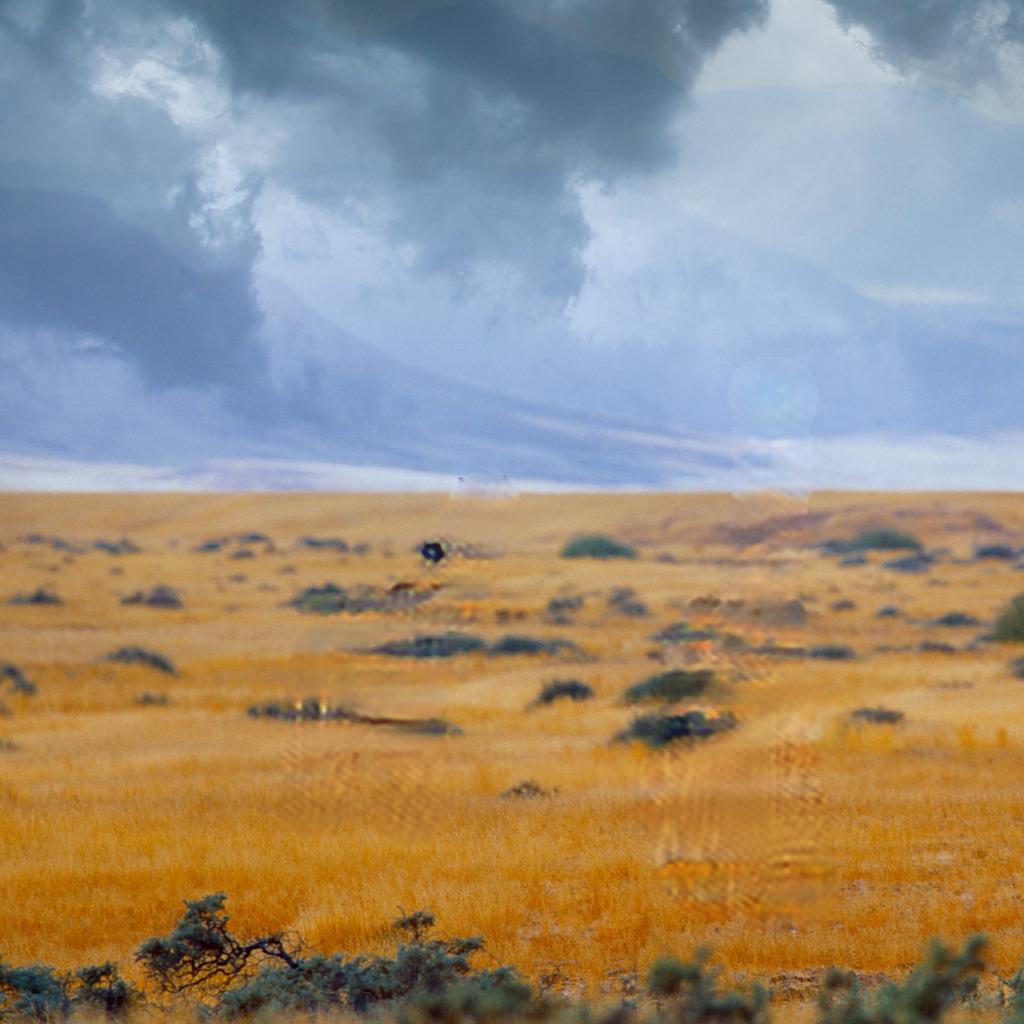}\\
		\includegraphics[width=0.32\linewidth]{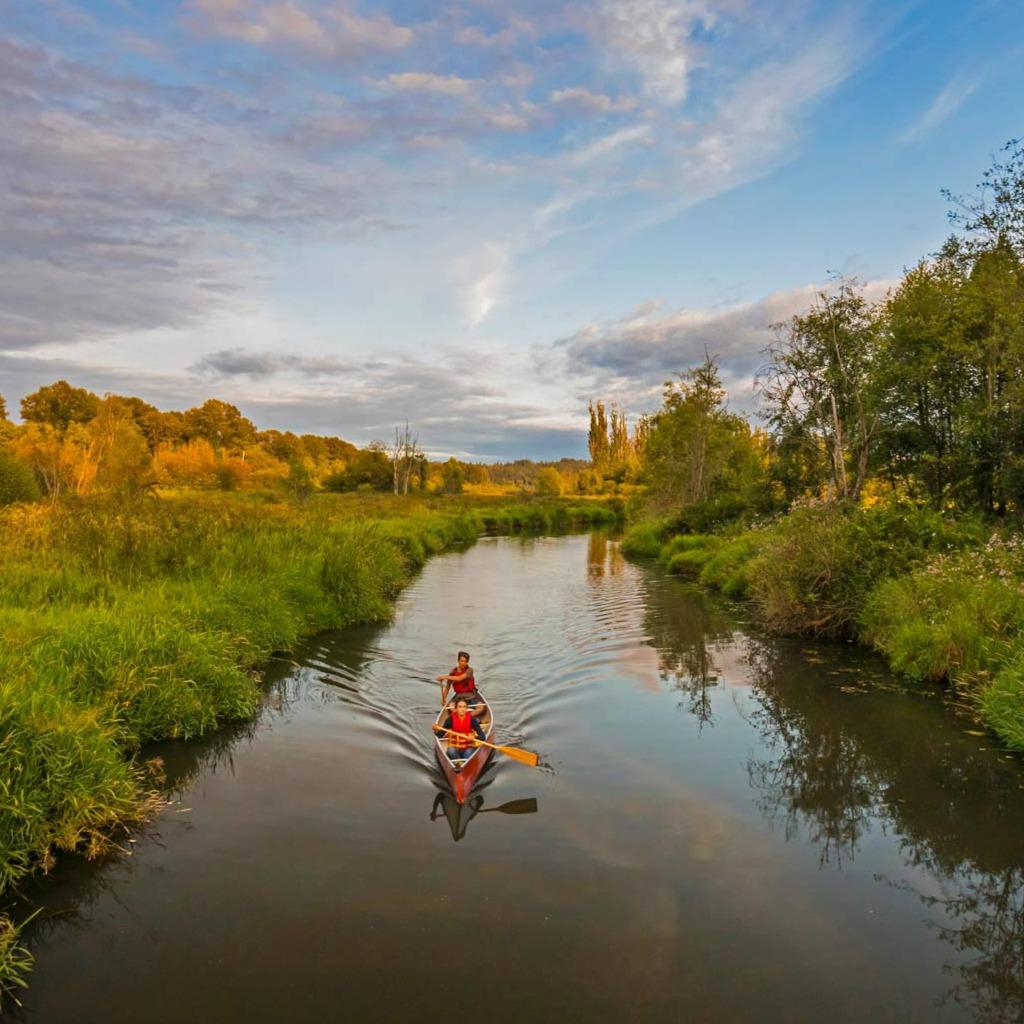}&
		\includegraphics[width=0.32\linewidth]{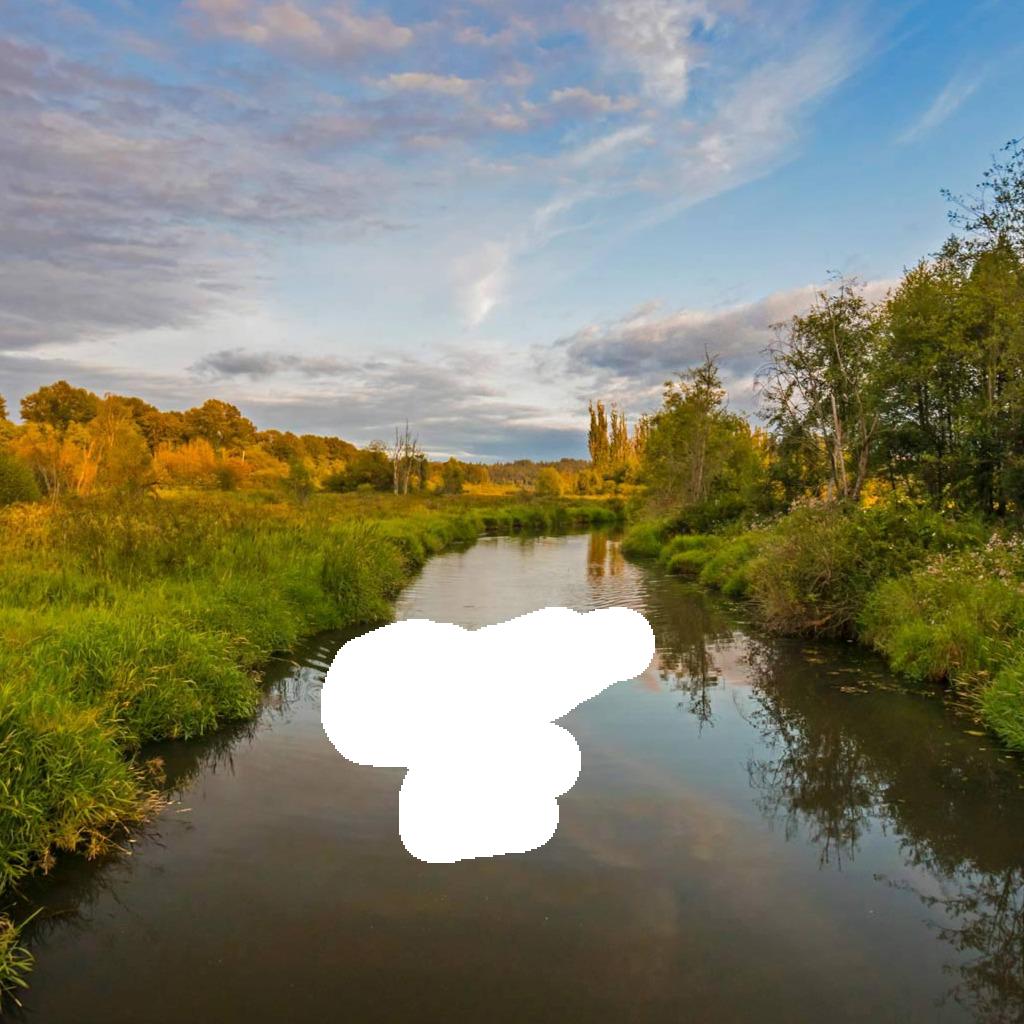}&
		\includegraphics[width=0.32\linewidth]{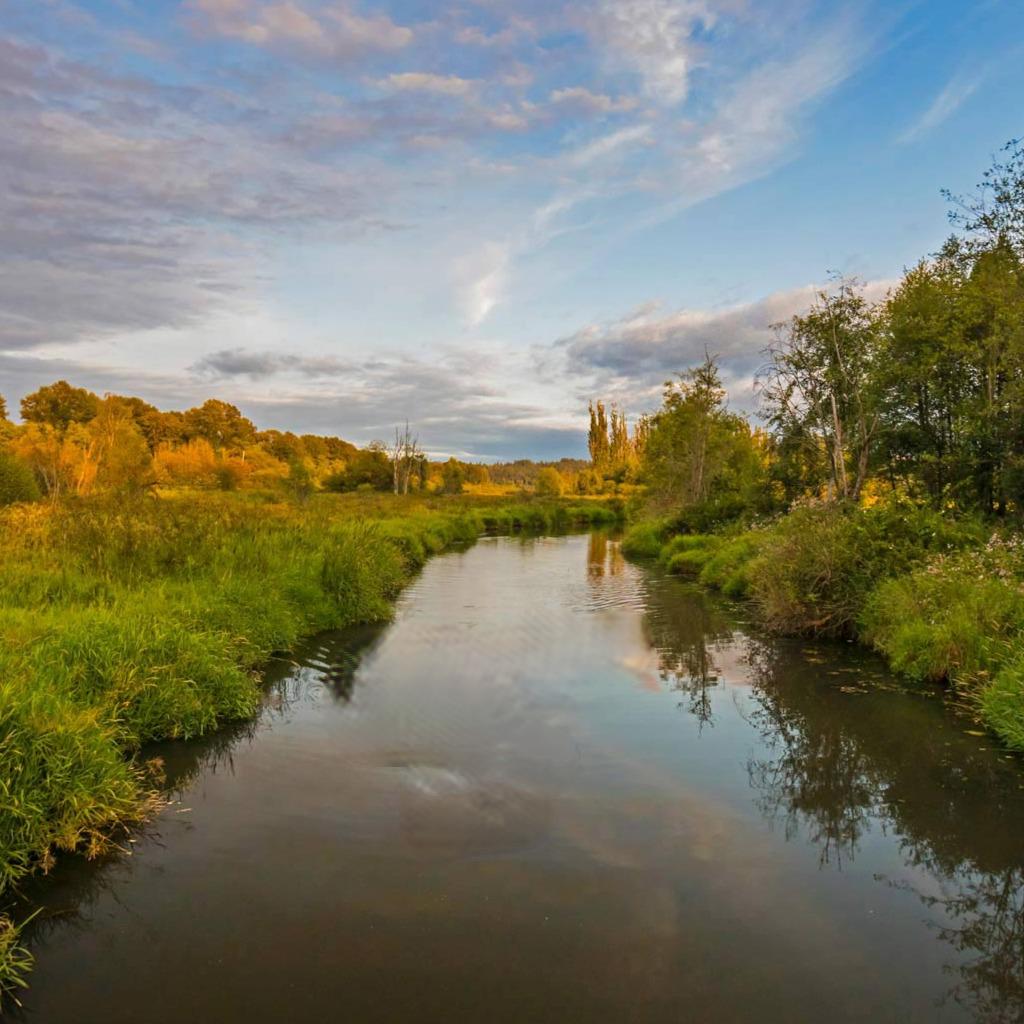}\\
		Original & Input & Ours			
	\end{tabular}
	\vspace{0.1in}
	\caption{Inpainting results on 1024$\times$1024 images using our PyramidFill. Zoom-in to see the details. The original images at the first two rows are extracted from CelebA-HQ dataset, and the original images at the last two rows are extracted from NSHQ dataset.}
	\label{Application}
\end{figure}

\subsection{Real Applications}
We study real use cases of image inpainting on high-resolution images using our PyramidFill, e.g., freckles removal, face editing, watermark removal, and general object removal in natural scenery. In the first row of Figure \ref{Application}, our model successfully removes the freckles on the face, as well as changing eyes. In the second row, the watermarks on the original image are removed successfully. In the third row, we remove the giraffes on the grassland and recover the background. In the last row, a boat is removed from the river. All inpainting results are realistic and with fine-grained details. Furthermore, our PyramidFill can also be used for more real applications, e.g., removing wrinkles on faces, changing hairstyles, restoring old photos.

\begin{table}[ht]
	\centering
	\caption{Ablation study on CelebA-HQ dataset with input size 64$\times$64 resolution, where the inputs are with center hole regions. Consistency regularization loss can improve the performance of our model.}
	\label{Quantitative5}
	\vspace{0.1in}
	\begin{tabular}
		{c|c|c|c}
		\hline
		      Methods  & L1 Loss$\downarrow$ & PSNR$\uparrow$ & SSIM$\uparrow$  \\
		\hline
		 Ours without $\mathcal{L}^{cons}_{D_{0}}$ & 0.0209  & 27.21 & 0.9129 \\
 		 Ours         & \textbf{0.0201}  & \textbf{27.54} & \textbf{0.9172}  \\
		\hline
	\end{tabular}
\end{table}

\begin{figure}[t!]
	\centering
	\begin{tabular}
		{@{\hspace{0.0mm}}c@{\hspace{0.2mm}}c@{\hspace{0.2mm}}c@{\hspace{0.2mm}}c@{\hspace{0.0mm}}}
		\includegraphics[width=0.25\linewidth]{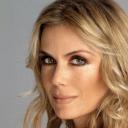}&
		\includegraphics[width=0.25\linewidth]{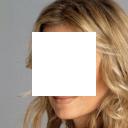}&
		\includegraphics[width=0.25\linewidth]{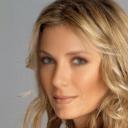}&
		\includegraphics[width=0.25\linewidth]{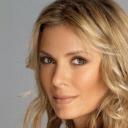}\\		
	\end{tabular}
	\caption{Ablation study of Refinement network in Texture Generators. From left to right, we show the real image, the masked input, the result with a one-stage network in Texture Generators and our result with a two-stage networks.}
	\label{ablation}
\end{figure}

\subsection{Ablation Study}
\noindent\textbf{Consistency regularization loss} To demonstrate the effects of consistency regularization loss used in the discriminator $D_0$, we retrain $\left\{G_0,D_0\right\}$ yet without consistency regularization loss on CelebA-HQ dataset masked with center holes. The results are provided in Table \ref{Quantitative5}, which shows that the consistency regularization loss can definitely improve the performance of PyramidFill on completing contents.

\noindent\textbf{Refinement network} In Texture Generators, a refinement network is designed as a second-stage for generating a finer result. We provide ablation experiments on CelebA-HQ dataset with 128$\times$128 images masked with center holes. For fair comparisons, two-stage networks are merged into a one-stage network, wherein the feature maps of the upsampling operator in the super-resolution network are directly inputted to the refinement network, and there is no coarse result to output. As shown in Figure \ref{ablation}, the two-stage network can generate a more photo-realistic inpainting result.
%------------------------------------------------------------------------
\section{Conclusions}
We present PyramidFill, a novel framework for the high-resolution image inpainting task. PyramidFill consists of a pyramid of PatchGANs, wherein the content GAN is responsible for generating contents in the lowest-resolution corrupted images, and each texture GAN is responsible for synthesizing textures at higher-resolution images progressively. We customized the generators and discriminators for the content GAN and texture GAN, respectively. Our model was trained on several datasets to evaluate its ability to fill correct contents and realistic textures for high-resolution image inpainting. Quantitative and qualitative results demonstrated the superiority of PyramidFill, comparing with several state-of-the-art methods.
%{\small
%\bibliographystyle{ieee_fullname}
%\bibliography{egpaper_final}
%}
%\input{egpaper_final.bbl}

\end{document}